\documentclass[10pt, conference, compsocconf, letterpaper]{IEEEtran}
\IEEEoverridecommandlockouts

\usepackage{cite}
\usepackage{amsmath,amssymb,amsfonts}
\usepackage{algorithm} 
\usepackage{algpseudocode} 
\usepackage{tikz}
\usepackage{graphicx}
\usepackage{textcomp}
\usepackage{subfig}
\usepackage{multirow}
 \def\BibTeX{{\rm B\kern-.05em{\sc i\kern-.025em b}\kern-.08em
    T\kern-.1667em\lower.7ex\hbox{E}\kern-.125emX}}
\usepackage{enumitem}
\usepackage{colortbl}
\usepackage{soul}
\usepackage{graphicx} 
\newlist{inlineroman}{enumerate*}{1}
\setlist[inlineroman]{itemjoin*={{, and }},afterlabel=~,label=\roman*.}

\newlist{Inlineroman}{enumerate*}{1}
\setlist[Inlineroman]{itemjoin*={{, and }},afterlabel=~,label=\Roman*.}

\begin{document}

\title{\text Autonomous Driving with Deep Reinforcement\\ Learning in CARLA Simulation}
 \author{\IEEEauthorblockN{Jumman Hossain}
 \IEEEauthorblockA{\textit{Department of Information Systems} \\
 \textit{University of Maryland, Baltimore County (UMBC)}\\
 jumman.hossain@umbc.edu}
 }


\maketitle

\begin{abstract}
Nowadays, autonomous vehicles are gaining traction due to their numerous potential applications in resolving a variety of other real-world challenges. However, developing autonomous vehicles need huge amount of training and testing before deploying it to real world. While the field of reinforcement learning (RL) has evolved into a powerful learning framework to the development of deep representation learning, and it is now capable of learning complicated policies in high-dimensional environments like in autonomous vehicles. In this regard, we make an effort, using Deep Q-Learning, to discover a method by which an autonomous car may maintain its lane at top speed while avoiding other vehicles. After that, we used CARLA simulation environment to test and verify our newly acquired policy based on the problem formulation.
\end{abstract}

\section{Introduction}
The development of learning-based methodologies, the proliferation of low-cost sensors, and the availability of enormous amounts of driving data have all contributed to the substantial progress that has been made in the field of autonomous driving over the course of the past few decades. In recent years, there has been an increase in the popularity of end-to-end approaches, which are methods that attempt to learn driving judgments directly from sensory inputs. This rise can be attributed to the rise of deep learning. Instead of first learning an exact representation of the data and then making judgments based on that information, the system will learn an intermediate representation in this fashion, which has the potential to produce superior results.

The navigation problem for AVs entails finding the best course of action to get the vehicle from one location to another without crashing into any of the moving obstacles or other vehicles along the way. Because AVs are supposed to maintain a safe distance from other vehicles while simultaneously making driving more fuel- and time-efficient, safety is a crucial part of navigation. Autonomous driving relies heavily on navigation, a topic that has been intensively researched in the fields of motion planning and mobile robotics.

In an autonomous navigation tasks, the goal of utilizing
a RL algorithm is to identify the best strategy for directing
the robot to its destination point through interaction with
the environment. Many well-known RL algorithms have been
adapted to create a RL-based navigation system, including DQN, DDPG, PPO, and their variants. These methods model
the navigation process as an MDP using sensor observation
as the state and the goal of maximizing the action’s predicted
revenue. RL-based navigation has the
advantages of being mapless, having a good learning ability,
and having a low reliance on sensor accuracy. Because RL is
a trial-and-error learning system, the physical training process
will ultimately result in the robot colliding with external
impediments.

Since autonomous driving is a situation in which an agent makes decisions based on what it senses, the problem can be turned into a Markov Decision Process (MDP), making it an ideal candidate for the application of Reinforcement Learning. While Deep Q Networks (DQNs) demonstrated superhuman performance in Atari games ~\cite{mnih2013playing}, ~\cite{mnih2015human}, and AlphaGo achieved widespread success ~\cite{godqn2016}, the use of Deep Reinforcement Learning for control-oriented tasks has experienced explosive growth.

The authors~\cite{Pyeatt1998LearningTR} make use of RL in order to get experience in racing in a virtual setting. There have also been several research published that demonstrate the application of inverse reinforcement learning (IRL) ~\cite{shen2022inverse, phan2022driving}to the task of autonomous driving. ~\cite{wulfmeier2016watch} Describe a system that can learn costmaps for autonomous driving utilizing IRL data directly from sensors. ~\cite{sharifzadeh2016learning} Demonstrate a strategy that would be used in real life in a straightforward case of highway driving on a specialized simulator.

In a manner analogous to that described above, we learn from unprocessed sensory data in a simulated setting. We make an effort to discover, through Deep Q-Learning, a method that would enable an autonomous car to maintain its lane at the highest potential speed while simultaneously avoiding collisions with other vehicles.

\section{Related Work}
\textbf{B-GAP}~\cite{mavrogiannis2020b} : Presented a navigation scheme that is based on deep reinforcement learning and considers driver behavior to perform behaviorally-guided navigation. Their DRL based policy implicitly models the interactions between traffic agents and computes safe trajectories for the ego-vehicle accounting for aggressive driver maneuvers such as overtaking, over-speeding, weaving, and sudden lane changes.

\textbf{Overtaking Maneuvers}~\cite{mehaOvertaking2018} : Presented a Deep Deterministic Policy Gradients (DDPG) based approach to learn overtaking maneuvers for a car, in presence of multiple other cars, in a simulated highway scenario. Their  training strategy resembles a curriculum learning approach that ables to learn smooth maneuvers, largely collision free, wherein the agent overtakes all other cars, independent of the track and number of cars in the scene.

\cite{humanDriverGraph} approaches for predicting the behaviors of human drivers based on the trajectories of their vehicles. They are using graph-theoretic tools to extract features from the trajectories and machine learning to learn a computational mapping between the trajectories and the driver behaviors. Their approach is similar to how humans implicitly interpret the behaviors of drivers on the road, by only observing the trajectories of their vehicles. By using machine learning to model this process, it may be possible to enable autonomous vehicles to better understand and predict the behaviors of human drivers, which can be useful for a variety of applications, including autonomous navigation.

In contrast, our objective is to implement Deep Q-Networks, which are capable of resolving the issue of driving autonomous vehicles in a highway environment as quickly as possible while simultaneously avoiding crashes.
\section{Preliminaries}

\subsection{Reinforcement Learning}
\begin{figure}[htbp]
\centerline{
\includegraphics[width=0.5\textwidth]{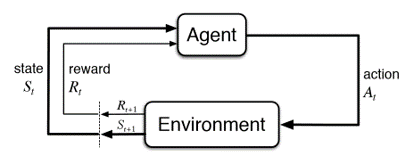}}
\caption{Agent-Environment Interaction in Reinforcement learning}
\label{rlfig}
\end{figure}

There are two primary components that fall under the umbrella of generalized reinforcement learning: an agent and an environment. At a certain moment in time $t$, the agent lies in a state $s_t$ of the environment; it then takes an action, moving from the state $s_t$ to a new state $s_{t+1}$, and receives a reward $r_t$ from the environment. The goal of the agent is to learn a policy $\pi$ that maximizes the cumulative reward over time. The standard agent-environment interaction in reinforcement learning ~\cite{sutton2018reinforcement} is depicted in Figure~\ref{rlfig}.

Typically, we can model an RL problem with decisions by formulating it as a Markov Decision Process (MDP). An MDP is often described as a tuple consisting of:

\begin{itemize}
  \item A state space S: which is the set of spaces in which an agent can find
himself along the process.
  \item An action state A: which is the set of actions available for the agent.
  \item A transition probability function that represents the probability to move from one state to another by choosing a certain action.
  \item A reward function $R_t(s_t, A_t)$ which is the reward obtained by choosing a certain action in a certain state.
\end{itemize}

In reinforcement learning, an agent interacts with an environment in a sequence of steps, with the goal of maximizing a reward signal. At each time step, the agent receives an observation of the environment and takes action based on this observation. The action leads to a new observation and a reward, and the process repeats.

The goal of the agent is to learn a policy that maps observations to actions in a way that maximizes the expected sum of rewards over time. This can be formalized using the following components:

\textbf{State:} A state is a representation of the environment at a given time step. It can include information about the agent's current location, the objects and obstacles in the environment, and any other relevant information.

\textbf{Action:} An action is a choice made by the agent at each time step. It can be a discrete decision, such as moving left or right, or a continuous decision, such as adjusting the speed of a robot.

\textbf{Reward:} A reward is a scalar value that is provided to the agent after each action. It reflects the quality of the action taken by the agent and is used to guide the learning process.

\textbf{Policy:} A policy is a function that maps states to actions. It determines the action that the agent should take at each time step based on the current state. The goal of the agent is to learn a policy that maximizes the expected sum of rewards.

The process of reinforcement learning can be summarized by the following equation:

$$\pi^*(s) = \arg\max_\pi \mathbb{E}[\sum_{t=0}^\infty \gamma^t r_t | s_0 = s, \pi]$$

Here, $\pi^(s)$ is the optimal policy for a given state $s$, $\pi$ is a policy being evaluated, $\mathbb{E}[\cdot]$ is the expected value operator, $\gamma$ is a discount factor that determines the importance of future rewards, and $r_t$ is the reward received at time step $t$. The sum is over all time steps starting at $t=0$ and going to infinity. The goal is to find the policy $\pi^(s)$ that maximizes the expected sum of rewards.

\subsection{Q-learning}

The mind of the agent in Q-learning is a table with the rows as the State or Observation of the agent from the environment and the columns as the actions to take. Each of the cells of the table will be filled with a value called Q-value which is the value that an action brings considering the state it is in. let’s call this table Q-table. The Q-table is actually the brain of the agent.

The agent starts taking an action in the environment and starts a Q-table initialized with zeros in all the cells. Then, the agent gets to a new state or observation (state is the information of the environment that an agent is in and observation is an actual image that the agent sees. Depending on the environment an agent gets state or observation as an input) by doing an action from the table. The state is a row in the table containing Q-values for each action and the highest value in the row will be the action that the agent takes (the column with the highest value in that specific row).

\begin{figure}[htbp]
\centering
\includegraphics[width=0.6\linewidth]{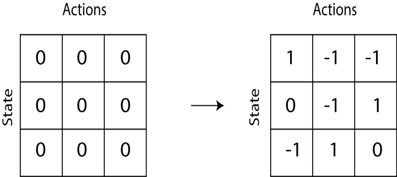}

\caption{Q-table.}
\end{figure}

In Q-learning, the expected future reward for a given action at a given time step is referred to as the action-value function or Q-value. The Q-value for an action at a time step is calculated using the following formula:

$$Q(s_t, a_t) = \mathbb{E}[R_{t+1} + \gamma \max_{a} Q(s_{t+1}, a) | s_t, a_t]$$

Here, $Q(s_t, a_t)$ is the Q-value for taking action $a_t$ at state $s_t$, $\mathbb{E}[\cdot]$ is the expected value operator, $R_{t+1}$ is the reward received at the next time step, $\gamma$ is the discount factor that determines the importance of future rewards, and $s_{t+1}$ is the state at the next time step. The term $\max_{a} Q(s_{t+1}, a)$ is the maximum Q-value over all actions at the next time step.

The Q-learning algorithm involves iteratively updating the Q-values using the above formula and a learning rate parameter $\alpha$. The Q-values are initially set to a random value and are updated using the following update rule:

$$Q(s_t, a_t) \leftarrow (1 - \alpha)Q(s_t, a_t) + \alpha (R_{t+1} + \gamma \max_{a} Q(s_{t+1}, a))$$

This update rule involves a combination of the current Q-value and the new estimate of the Q-value, with the learning rate $\alpha$ determining the relative importance of these two terms. The Q-learning algorithm is run for a fixed number of iterations or until the Q-values converge to a stable value.

Once the Q-learning algorithm has completed, the optimal policy can be obtained by choosing the action with the highest Q-value at each state. The Q-table is a data structure that stores the Q-values for all states and actions, and can be used to look up the optimal action for a given state.

The agent gets a reward by doing the action. The reward has a meaning. the higher the value the better, but sometimes the lower value could mean that the agent has taken the better action. The reward comes from the environment and the environment defines which of the lower or higher reward is better. The environment gives the agent the reward for taking an action in a specific state.

The agent keeps doing steps 1 to 3 and gathers information in its “memory”. The memory contains tuples of state, next state, action, reward, and a Boolean value for indicating the termination of the agent. These steps keep on going and the agent memorizes the info until the termination happens.

After the termination of the agent which could mean completing the task or failing, the agent starts replaying the experiences it gathered in its memory. A batch of a particular size will be chosen from the memory and the task of training will be performed on it. Basically, this means that the Q-table starts filling up. This is called Experience Replay

Basically, Q-value is the reward obtained from the current state plus the maximum Q-value from the next state. So, that means the agent has to get the reward and the next state from its experience in the memory and add the reward to the highest Q-value derived from the row of the next state in the Q-table and the result will go into the row of the current state and the column of the action, both obtained from the experience in the memory.

\subsection{Deep Q-Learning}
The only difference between Q-learning and DQN is the agent’s brain. The agent’s brain in Q-learning is the Q-table, but in DQN the agent’s brain is a deep neural network.

 DQN uses a neural network (Fig. 3) to represent the value function, which is trained using a variant of the Q-learning algorithm. The input to the neural network is the state or observation, and the output is the Q-value for each possible action. The neural network is trained using a loss function that measures the difference between the predicted Q-values and the target Q-values, which are updated using the Bellman equation.

Overall, the main difference between Q-learning and DQN is the way that the value function is represented and updated. Q-learning uses a tabular representation, while DQN uses a neural network representation. Both algorithms can be effective for learning policies in different types of environments, but DQN may be better suited for problems with large or continuous action and observation spaces. Taking all of this into account, DQN might be an appropriate option for an autonomous driving situation which includes: 

\textbf{Real-time performance:} DQN can learn policies in real-time, which is important for applications like autonomous driving where decisions need to be made quickly and accurately.

\textbf{Generalization:} DQN can learn policies that generalize well to different scenarios and environments, which is important for autonomous driving, where the vehicle may need to operate in a range of different conditions.

\textbf{Scalability:} DQN can scale to large action and observation spaces, which is often the case in autonomous driving, where the vehicle may need to consider a wide range of possible actions and observations.

\textbf{Robustness:} DQN can learn robust policies that are resistant to noise and uncertainty, which is important for autonomous driving, where the environment may be highly dynamic and unpredictable.

\begin{figure}[htbp]
\centering
\includegraphics[width=0.8\linewidth]{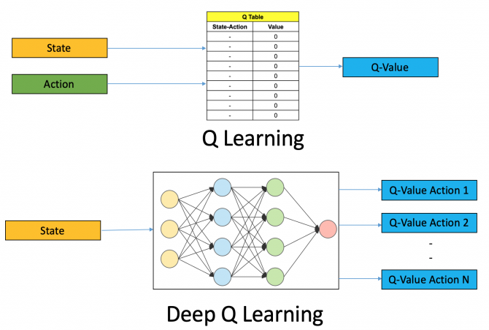}

\caption{Q-learning vs Deep Q-learning.}
\end{figure}

\section{Problem Formulation}

We formulate the problem with observation space, action space, and reward structure throughout the environment interactions by agents.

\textbf {Observation Space:} The Observation space for each agent is a 5 by 5 array that has a list of five vehicles that are located in close proximity to one another and is described by a set of characteristics including Position (x, y) and Velocity (Vx, Vy). In order to accomplish the goals of multi-agent learning, we create a tuple by concatenating the observation of the number of agents that are now active.

\textbf {Action Space:} Our Action Space is a Discrete space that consists of Lateral movement and Longitudinal movement. 

        $0: LANE LEFT$,

        $1: IDLE$,

        $2: LANE RIGHT$,

        $3: FASTER$,

        $4: SLOWER$

\textbf{Reward Formulation:} On the highway, which is quite straight and has numerous lanes, the driver of the vehicle receives a reward for maintaining a high speed, staying in the lanes to the right of the center, and avoiding collisions.

$RIGHT LANE REWARD = 0.1$

$HIGH SPEED REWARD = 0.4$

$LANE CHANGE REWARD = 0$

Within our learning policy, we take the mean of the reward values that were acquired by each agent. The episode is over when it reaches a maximum of 50 seconds in length or when the agents come into physical contact with one another.
\begin{table*}[t]
\centering
\begin{tabular}{ccc}
\begin{tabular}[t]{|l|ll|ll|}
\hline
\multicolumn{1}{|c|}{\multirow{2}{*}{Vehicle}} & \multicolumn{2}{c|}{Position} & \multicolumn{2}{c|}{Velocity} \\ \cline{2-5} 
\multicolumn{1}{|c|}{} & \multicolumn{1}{c|}{X} & Y & \multicolumn{1}{c|}{Vx} & Vy \\ \hline
Ego-Vehicle & 0.04 & 0.05 & 0.72 & 0 \\ \hline
Vehicle 1 & -0.12 & 0 & -0.16 & 0 \\ \hline
Vehicle 2 & 0.09 & 0.03 & -0.072 & 0 \\ \hline
Vehicle 3 & 0.06 & 0.02 & -0.65 & 0 \\ \hline
Vehicle 4 & 0.28 & 0.18 & 0.20 & 0.025 \\ \hline
\end{tabular} & \begin{tabular}[t]{|l|l|}
\hline
\multicolumn{1}{|c|}{Value} & \multicolumn{1}{c|}{Action} \\ \hline
0 & LANE LEFT \\ \hline
1 & IDLE \\ \hline
2 & LANE RIGHT \\ \hline
3 & FASTER \\ \hline
4 & SLOWER \\ \hline
\end{tabular} & \begin{tabular}[t]{|l|l|}
\hline
\multicolumn{1}{|c|}{Action} & Reward \\ \hline
RIGHT LANE & 0.1 \\ \hline
HIGH SPEED & 0.4 \\ \hline
LANE CHANGE & 0 \\ \hline
\end{tabular} \\
(a) Observation Space & (b) Action Space & (c) Reward Space
\end{tabular}
\caption{Problem formulation with (a) observation space: Position and velocity of different vehicles, (b): Action space with 5 scenarios, and (c) reward values of three actions.}
\label{tab:problem-formulation}
\end{table*}


\section{Methodology}
\label{method}

In a multi-agent problem with partially observed data, it can be challenging to design effective reinforcement learning algorithms that can make long-term strategies over thousands of steps in a large action space. One approach that may be useful in this scenario is to use multi-agent reinforcement learning algorithms.

Multi-agent reinforcement learning algorithms are designed specifically to address problems involving multiple agents interacting with each other and the environment. These algorithms typically involve the use of various techniques to handle the complexity of multi-agent systems, such as decentralized control, cooperative and competitive behavior, and communication between agents.

There are several different approaches to multi-agent reinforcement learning, including centralised training with decentralised execution, independent learning, and cooperative learning. Each of these approaches has its own benefits and limitations, and the choice of which approach to use will depend on the specific characteristics of the problem being addressed.

It is also important to note that in order to effectively solve a multi-agent problem with partially observed data, it may be necessary to incorporate additional techniques such as partially observable Markov decision processes (POMDPs) or belief state representations. These techniques can help the agents to reason about the uncertainty in their observations and make more informed decisions.

In the Q-learning algorithm, the action-value function Q(s, a) is used to estimate the expected future reward of taking a particular action a in a particular state s, and to select the action that will maximize this expected reward. This expected reward is also known as the "utility" of the action, and it takes into account the immediate reward of taking the action as well as the future rewards that may be obtained by following a particular policy.

The action-value function Q(s, a) is typically represented as a table or a function approximator, such as a neural network. It is updated using the Bellman equation, which expresses the relationship between the expected utility of an action and the expected utility of the subsequent state that the action leads to. The Bellman equation is used to update the action-value function Q(s, a) in an iterative fashion, as the agent explores the environment and learns about the consequences of its actions.

In the example you provided of a chess game, the action-value function Q(s, a) could be used to measure how beneficial it is to move a particular pawn forward in a particular game state. The action-value function would take into account the immediate reward of making the move (e.g., capturing an opponent's piece) as well as the potential future rewards that may be obtained by following a particular strategy (e.g., setting up a winning position).  

For the sake of this project, the Q assesses how well one is able to decide whether to speed up, slow down, or switch lanes at any given time and in any given setting. As the agent examines the present state of the environment and chooses an action, the environment transitions into a new state while simultaneously returning a reward to signal the result of the action that was chosen. Because of Q-learning, the agent is equipped with a cheat sheet that details the Q-values present in each scenario along with the actions that should be taken in response to them. On the other hand, the Q-learning agent does not have the ability to make value estimates for states that have not yet been encountered. If the Q-learning agent has never been exposed to a state before, it will be completely clueless regarding the appropriate next steps to take.

\begin{figure}[htbp]
\centering
\includegraphics[width=0.35\linewidth]{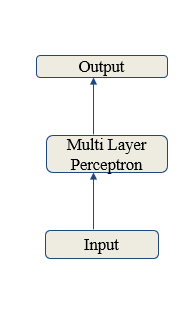}

\caption{Multi-Layer Perceptron Model.}
\vspace{-2ex}
\end{figure}

In order to acquire knowledge about Q values via a Multi-Layer Perceptron Model, we make use of a DQN technique. The actions are selected by the policy based on which ones have the highest possible Q value for the current condition. While the DQN algorithm works by using a neural network to predict the expected reward for taking a particular action in a given state. The neural network is trained using a variant of stochastic gradient descent called Q-learning, which updates the network's weights based on the difference between the predicted reward and the actual reward received after taking the action. We employ the DQN method to learn Q values through a Multi-Layer Perceptron Model. The actions are chosen by the policy with maximum Q value for a given state.

\texttt{RIGHT\_LANE\_REWARD}, \texttt{HIGH\_SPEED\_REWARD}, and \texttt{LANE\_CHANGE\_REWARD} are variables that define the reward values for certain actions taken by the autonomous vehicle (AV) during training.

\texttt{RIGHT\_LANE\_REWARD} is a positive value that rewards the AV for staying on the rightmost lanes. This reward encourages the AV to drive in a safe and efficient manner.

\texttt{HIGH\_SPEED\_REWARD} is a positive value that rewards the AV for driving at a high speed. This reward encourages the AV to drive efficiently and reach its destination quickly.

\texttt{LANE\_CHANGE\_REWARD} is a value that rewards or penalizes the AV for changing lanes. If \texttt{LANE\_CHANGE\_REWARD} is positive, it will encourage the AV to change lanes. If \texttt{LANE\_CHANGE\_REWARD} is negative, it will discourage the AV from changing lanes. If \texttt{LANE\_CHANGE\_REWARD} is set to zero, the AV will not be specifically rewarded or penalized for lane changes.

These reward values are used in the reward formulation of the DQN algorithm \ref{alg:dqn} to train the AV to perform a particular task, such as driving on a highway. The AV will receive rewards or penalties based on the values of \texttt{RIGHT\_LANE\_REWARD}, \texttt{HIGH\_SPEED\_REWARD}, and \texttt{LANE\_CHANGE\_REWARD} after taking actions in different states, and will use this information to learn the best actions to take in different situations.

In this formulation, the neural network $Q$ is used to approximate the optimal action-value function, which determines the best action to take in a given state. The replay memory $D$ is used to store transitions and sample minibatches for training the network. The $\epsilon$-greedy policy is used to balance exploration (trying new actions) and exploitation (choosing the action with the highest predicted reward). The discount factor $\gamma$ determines the importance of future rewards in the action-value function. And the loss function $L$ is used to update the weights of the neural network to better approximate the optimal action-value function.

\begin{algorithm}[H]
\caption{DQN Algorithm for AV Driving}
\label{alg:dqn}
\begin{algorithmic}[1]
\State Initialize the neural network with random weights: $W$.
\State Initialize the replay memory $D$ to capacity $N$.
\For{each step $t = 1, 2, 3, \ldots$}
    \State Receive the current state $s$ of the AV.
    \State Select an action $a$ using an $\epsilon$-greedy policy based on the current estimate of the action-value function: $Q(s, a; W)$.
    \State Execute action $a$ in the environment.
    \State Observe the reward $r$ and the next state $s'$.
    \State Store the transition $(s, a, r, s')$ in the replay memory $D$.
    \State Sample a minibatch of transitions $(s_i, a_i, r_i, s'_i)$ from the replay memory $D$.
    \If{episode terminates at step $i+1$}
        \State $y_i \gets r_i$
    \Else
        \State $y_i \gets r_i + \gamma \max_{a'} Q(s'_i, a'; W)$
    \EndIf
    \State Update the weights of the neural network using gradient descent to minimize the loss function:
    \State $L_i(W) \gets (y_i - Q(s_i, a_i; W))^2$
\EndFor
\end{algorithmic}
\end{algorithm}

\begin{table*}[t]
\centering
\begin{tabular}{|l|l|l|l|l|l|l|l|l|l|l|}
\hline
Metric/Town    &                                               & Town01                                              & Town02                                              & Town03                                              & Town04                                              & Town05                                              & Town06                                              & Town07                                              & Town10                                              & Total                                                         \\ \hline
Collision rate & \begin{tabular}[c]{@{}l@{}}S\\ U\end{tabular} & \begin{tabular}[c]{@{}l@{}}0.78\\ 0.98\end{tabular} & \begin{tabular}[c]{@{}l@{}}0.85\\ 0.98\end{tabular} & \begin{tabular}[c]{@{}l@{}}0.72\\ 0.99\end{tabular} & \begin{tabular}[c]{@{}l@{}}0.81\\ 0.99\end{tabular} & \begin{tabular}[c]{@{}l@{}}0.38\\ 0.95\end{tabular} & \begin{tabular}[c]{@{}l@{}}0.4\\ 0.92\end{tabular}  & \begin{tabular}[c]{@{}l@{}}0.77\\ 0.89\end{tabular} & \begin{tabular}[c]{@{}l@{}}0.58\\ 0.90\end{tabular} & \begin{tabular}[c]{@{}l@{}}68\%\\ 96\%\end{tabular}           \\ \hline
Speed          & \begin{tabular}[c]{@{}l@{}}S\\ U\end{tabular} & \begin{tabular}[c]{@{}l@{}}8.60\\ 5.98\end{tabular} & \begin{tabular}[c]{@{}l@{}}8.22\\ 5.7\end{tabular}  & \begin{tabular}[c]{@{}l@{}}8.43\\ 5.99\end{tabular} & \begin{tabular}[c]{@{}l@{}}9.05\\ 6.02\end{tabular} & \begin{tabular}[c]{@{}l@{}}9.26\\ 6.25\end{tabular} & \begin{tabular}[c]{@{}l@{}}9.23\\ 6.45\end{tabular} & \begin{tabular}[c]{@{}l@{}}7.54\\ 5.68\end{tabular} & \begin{tabular}[c]{@{}l@{}}8.64\\ 6.31\end{tabular} & \begin{tabular}[c]{@{}l@{}}8.71 km/h\\ 6.10 km/h\end{tabular} \\ \hline
Timesteps      & \begin{tabular}[c]{@{}l@{}}S\\ U\end{tabular} & \begin{tabular}[c]{@{}l@{}}315\\ 190\end{tabular}   & \begin{tabular}[c]{@{}l@{}}329\\ 208\end{tabular}   & \begin{tabular}[c]{@{}l@{}}365\\ 235\end{tabular}   & \begin{tabular}[c]{@{}l@{}}371\\ 209\end{tabular}   & \begin{tabular}[c]{@{}l@{}}420\\ 250\end{tabular}   & \begin{tabular}[c]{@{}l@{}}473\\ 323\end{tabular}   & \begin{tabular}[c]{@{}l@{}}272\\ 214\end{tabular}   & \begin{tabular}[c]{@{}l@{}}341\\ 270\end{tabular}   & \begin{tabular}[c]{@{}l@{}}373\\ 241\end{tabular}             \\ \hline
Total Reward   & \begin{tabular}[c]{@{}l@{}}S\\ U\end{tabular} & \begin{tabular}[c]{@{}l@{}}2165\\ 510\end{tabular}  & \begin{tabular}[c]{@{}l@{}}2026\\ 498\end{tabular}  & \begin{tabular}[c]{@{}l@{}}1995\\ 591\end{tabular}  & \begin{tabular}[c]{@{}l@{}}1789\\ 562\end{tabular}  & \begin{tabular}[c]{@{}l@{}}2188\\ 486\end{tabular}  & \begin{tabular}[c]{@{}l@{}}2010\\ 694\end{tabular}  & \begin{tabular}[c]{@{}l@{}}1485\\ 354\end{tabular}  & \begin{tabular}[c]{@{}l@{}}1902\\ 514\end{tabular}  & \begin{tabular}[c]{@{}l@{}}1954\\ 513\end{tabular}            \\ \hline
\end{tabular}
\caption{Performance of our agents: standard (S), and untrained (U).The
results have been aggregated over the two weather sets (soft and hard), and three traffic scenarios (no, regular, and dense).}
\end{table*}

\begin{figure}[htbp]
\centering
\includegraphics[width=0.9\linewidth]{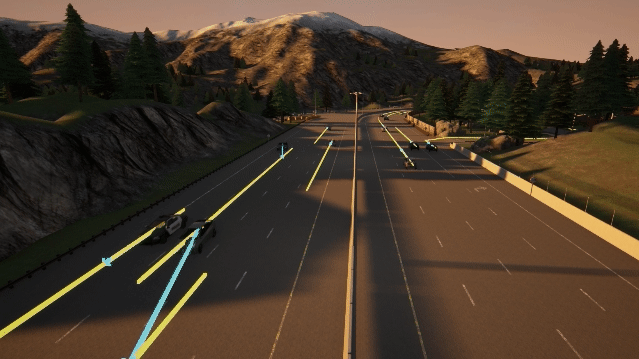}

\caption{CARLA Highway Environment (Town04).}
\end{figure}

\section{Results}
\label{results}
Across all of CARLA~\cite{dosovitskiy2017carla}'s towns, under a variety of climatic circumstances, and in three distinct traffic patterns, we put our agent through a rigorous test, evaluating it based on four key parameters.
\begin{itemize}
\item \textbf{Metrics:} collision rate, 
speed, total reward, and timesteps (i.e. the number of
steps without any infraction or collision).
\item \textbf{Towns:} Every town in CARLA has certain distinguishing characteristics all its own. Town04 (Fig. 5) served as the platform for our agent's training, and subsequent towns—including Town01, Town02, Town03, Town04, Town05, Town06, Town07, and Town10—were used to assess its performance.

\item \textbf{Weather:} Our evaluation is based on two separate sets of weather presets: easy and difficult. The first set, which is just utilized for training , and the second set, which is completely new to the agent, are as follows: WetCloudyNoon, CloudySunset, WetCloudySunset, HardRainNoon.

\item \textbf{Traffic:}  Our agent is evaluated using three distinct types of traffic: no traffic, regular traffic, and dense traffic.

\end{itemize}
The advantages of DQN are also measured by contrasting the performance of the same agent with and without DQN; the former is denoted by the letter "S" for Standard DQN, while the latter is denoted by the letter "U" for the baseline performance of an agent with the same architecture and fixed random weights throughout the evaluation. Table 2 demonstrates our agents' aggregate performance across both traffic scenarios and both weather sets.

The formulation of the method has the ability to maximize the standard reward that is gained by all of the agents. It's not uncommon for the reward signal to fluctuate or exhibit non-monotonic behavior as the agent learns, and it can take some time for the reward to converge to a stable value or trend. It appears that after 3000 steps, the reward begins to converge with an upward trend. This suggests that the algorithm is learning a policy that is able to consistently achieve a high reward.

\begin{figure}[htbp]
\centering
\includegraphics[width=0.8\linewidth]{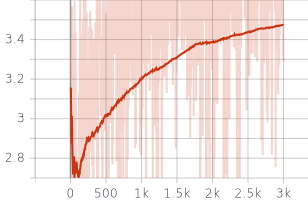}

\caption{Policy learning with DQN.}
\end{figure}

\section{Limitations}

There are several potential limitations of the proposed DQN approach. 

\textbf{Sensors and perception:} Depending on the sensors and perception capabilities of the agents, it has limited visibility or understanding of the environment, which could affect the ability to make informed decisions.

\textbf{Reward signal:} The reward signal used in the simulation (i.e., the speed of the vehicles within the range of [20, 30]) may not capture all of the relevant factors that influence the agents' performance, leading to suboptimal policies.

\textbf{Limited generalizability:} The policies learned by the agents in this simulation may not generalize well to other environments or scenarios, meaning that they may not perform well when applied to different situations.

\section{Conclusion and Future Work}
 We have used reinforcement learning to learn a policy with a modified DQN algorithm to train autonomous vehicles to achieve maximum speed while avoiding collisions. We also designed our own observation, action, and reward structure. We have experimented with this problem formulation in CARLA simulation environment, and considering future work that involves applying other reinforcement learning algorithms, such as Deep Deterministic Policy Gradient (DDPG)~\cite{lillicrap2015continuous}, Soft Actor-Critic (SAC)~\cite{haarnoja2018soft}, and possibly training the policy in a decentralized manner using Proximal Policy Optimization (PPO)~\cite{schulman2017proximal}.

\bibliographystyle{unsrt}
\bibliography{bibliography}
\end{document}